\documentclass[a4paper]{article}

\usepackage{INTERSPEECH_v2}
\usepackage{diagbox}
\usepackage{array}
\usepackage{hhline}
\usepackage{xcolor}
\usepackage{colortbl}
\usepackage{stfloats}
\title{Towards Speech Emotion Recognition ``in the wild'' \\using Aggregated Corpora and Deep Multi-Task Learning}
\name{Jaebok Kim, Gwenn Englebienne, Khiet P. Truong, Vanessa Evers}
\address{
  Human Media Interaction, University of Twente, Enschede, The Netherlands
  }
\email{\{j.kim, g.englebienne, k.p.truong, v.evers\}@utwente.nl}

\begin{document}

\maketitle
\begin{abstract}
One of the challenges in Speech Emotion Recognition (SER) ``in the wild'' is the large mismatch between training and test data (e.g. speakers and tasks). In order to improve the generalisation capabilities of the emotion models, we propose to use Multi-Task Learning (MTL) and use gender and naturalness as auxiliary tasks in deep neural networks. This method was evaluated in within-corpus and various cross-corpus classification experiments that simulate conditions ``in the wild''. In comparison to Single-Task Learning (STL) based state of the art methods, we found that our MTL method proposed improved performance significantly. Particularly, models using both gender and naturalness achieved more gains than those using either gender or naturalness separately. This benefit was also found in the high-level representations of the feature space, obtained from our method proposed, where discriminative emotional clusters could be observed.



\end{abstract}
\noindent\textbf{Index Terms}: speech emotion recognition, computational paralinguistics, deep learning

\section{Introduction}

Due to the increasing availability of crowd platforms and smart devices, it becomes realistic to collect an enormous amount of speech data from large and diverse populations. With deep learning technology, it is feasible to build a reliable model that captures more abstract properties of large corpora for Speech Emotion Recognition (SER). However, the performance in the wild is still unreliable partly because of diversity and often unknown contextual factors such as tasks, speakers, and recording conditions that can be encountered in the wild. \cite{schuller2010cross,ververidis2006emotional}. The diversity is not always covered in the corpora available to the researcher. Aggregating the corpora in the hope that the model obtained is general and robust enough against these differences has yielded variable results. Currently, speech emotion models work best if they are applied under circumstances that are similar to the ones that the model was trained on \cite{schuller2010cross}. 
Normalisation~\cite{schuller2010cross}, selection of samples~\cite{schuller2011selecting}, and transfer learning or adaptation~\cite{zhang2011unsupervised,deng2014autoencoder,deng2014linked} have been studied as methods. It was found that cross-corpus training worked to a certain degree only if corpora have similar contexts. To improve the generalisability of speech emotion models, we propose to use multi-task learning (MTL) that takes the differences between corpora into account as subtasks (i.e. gender and naturalness) and learns from these to make better classifications ``in the wild'' conditions.

MTL finds a common and essential representation between different tasks and often improves generalisation of a main task \cite{caruna1993multitask}. However, the success of MTL heavily depends on the choice of subtasks that may (not) relate to the main task. 
MTL has been applied in computer vision tasks~\cite{jou2016deep}, but for SER tasks, the use of MTL is relatively new. In the field of SER, naturalness~\cite{vogt2005comparing} and speaker characteristics~\cite{brody1985gender} such as gender and age, affect the way emotions are expressed, and they are commonly accessible from meta information of emotional speech corpora. However, it is unknown if they are helpful as subtasks in MTL. 

In this paper, we investigated whether MTL using gender and naturalness as subtasks improves generalisability of the emotion models trained. Compared to previous studies using deep learning, we validated our method using not only within-corpus but also cross-corpus settings reflecting more realistic challenges. In contrast to other commonly used methods such as transfer learning that requires additional data from test corpora~\cite{zhang2011unsupervised,deng2014autoencoder,deng2014linked}, our method aims to generalise the training model without having access to test data which resembles ``the wild setting''. To the best of our knowledge, this is the first work that investigates gender and naturalness as subtasks of deep MTL in order to model large and variable emotional speech samples.

This paper is structured as follows. We first introduce related previous work in Section~\ref{sec:relatedwork}. Especially, we give a short overview of MTL and high-level representation of emotional speech. Next, we present corpora in Section~\ref{sec:data}, and describe our proposed learning scheme in Section \ref{sec:method}. The results will be reported in Section~\ref{sec:result} and concluded in Section~\ref{sec:conclusion}.

\section{Related Work}\label{sec:relatedwork}

MTL is a machine learning approach that learns a main task with other related subtasks at the same time by using a shared representation \cite{baxter2000model,caruna1993multitask,caruana1998multitask}. MTL in deep neural networks (DNN) is similar to single-task learning except for the topology. It allows the learner to use the commonalities among the tasks, which often leads to improved generalisation. Hence, MTL is often regarded as a sort of inductive transfer. It improves generalisation by using the domain information extracted from the training signals of related tasks as inductive biases. 
However, 
choosing subtasks is a critical decision for MTL and helpful tasks for essential learning do not necessarily have to be related to the main task \cite{caruana1998multitask}. 
Recently, MTL has been applied to various fields such as computer vision and speech recognition \cite{seltzer2013multi,zhang2014facial,evgeniou2007multi}. Particularly, MTL with SVM was applied to SER in a cross corpus setting \cite{zhang2016cross}. Their method did not share hyperplanes but rather information to train separated hyperplanes simultaneously. In \cite{eyben2012multitask}, Long Short Term Memory (LSTM) based MTL was explored but limited to main tasks such as regression of arousal and valence, and confidence of annotations. In order to see the benefits of MTL, i.e. ability of generalisation, cross-corpus classification experiments simulating ``in the wild'' conditions should be carried out which have not been done before with MTL.

\begin{table*}[!htb]
\centering
\footnotesize{
\caption{Overview of the selected corpora (the number of utterances)}\label{tab:corpora}
\begin{tabular}{|l|r|r|r|r|r|r|r|r|r|r|}
\hline
Corpus (ID) & Speakers & \multicolumn{4}{|c|}{Emotion} & \multicolumn{2}{|c|}{Gender} & \multicolumn{2}{|c|}{Naturalness} & Languages\\ \cline{3-10} 
  & & neutral & happy & sad & angry & female & male & natural & acted& \\
\hline                
\hline
AIBO (A)    & 51 & 10967  & 889 & 0 & 1492  & 7579 & 5769 & 13348 & 0 & German \\
EMODB (E)   & 10 & 77 & 61  & 58  & 97 & 160 & 133 & 0 & 293  & German\\
ENTERFACE (F) & 43 & 0  & 208 & 422 & 211 & 200 & 641 & 0 & 841 & English\\
LDC (L)     &   7 & 80  & 180 & 161 & 139 & 320 & 240 & 0 & 560 & English\\
IEMOCAP (I)   &   10  & 1708  & 595 & 2168  & 2206 & 336 & 6341 & 3177 & 3500 & English\\
\hline
\hline
total     & 121 & 12832 & 1933  & 2809  & 4145  & 8595 & 13124 & 16525 & 5194 & \\
\hline
\end{tabular}
}
\end{table*}

As deep learning has gained a lot of interest and showed promising results in various fields of automatic speech analysis~\cite{mohamed2012acoustic}, it is being actively investigated in the field of SER too. Especially, by using representation learning~\cite{bengio2013representation}, there has been effort to extract unsupervised features which generalise emotional speech rather than engineered features (e.g. pitch) \cite{ghosh2015learning,trigeorgis2016adieu}. \cite{kunHan2014dnn,lee2015high} proposed an intrinsic way to build a high-level representation of emotion using the engineered features. They extracted segment-level engineered features (e.g. Mel-Frequency Cepstral Coefficients (MFCC) and pitch) and modelled probabilities of emotional categories using Deep Neural Network (DNN)~\cite{kunHan2014dnn} and Bi-directional LSTM (BLSTM)~\cite{lee2015high}. Then, functionals of the probabilities were used to extract utterance-level features, denoted as high-level feature representation. Extreme Learning Machine (ELM) using the utterance-level features outperformed conventional approaches such as HMM, SVM, and BLSTM. More recently, \cite{ghosh2015learning} showed that unsupervised representation learning has limitation in complex subsequent structures for affect compared to \cite{lee2015high}'s approach. Therefore, we chose \cite{kunHan2014dnn,lee2015high}'s architectures as baselines (single task learning) and compared it to our proposal and will investigate the effectiveness of MTL in various settings.

\section{Data}\label{sec:data}

In order to test our method with varying contexts, we used multiple corpora. 
We decided to focus on four representative emotional categories: neutral, happy, sad, and angry. We also aimed for variation in gender (female or male) and naturalness (natural or acted). We selected six corpora meeting our requirements: LDC Emotional Prosody (L)~\cite{liberman2002emotional}, eNTERFACE (F)~\cite{martin2006enterface}, EMODB (E)~\cite{burkhardt2005database} FAU-aibo emotion corpus (A)~\cite{batliner2004you}, and IEMOCAP (I)~\cite{busso2008iemocap}. 
Table \ref{tab:corpora} summarizes the corpora selected. IEMOCAP has both acted and natural (improvised) emotional speech~\cite{busso2008iemocap}, while EMODB, ENTERFACE, and LDC only have acted speech. Since AIBO has only emotional speech of children, we further categorised the gender of FAU-aibo corpus to female-child and male-child. While languages pose cultural
differences in emotional expressions  \cite{schuller2010cross}, we discard languages as a subtask since the selected corpora have only Germanic languages.

\section{Method}\label{sec:method}

\subsection{Features}
Based on previous work~\cite{kunHan2014dnn,lee2015high}, we chose a feature set: F0, voice probability, zero-crossing-rate, 12-dimensional MFCC with energy and their first time derivatives, totalling 32 features. As shown~\cite{kunHan2014dnn,lee2015high}, lower level features (e.g. mel-spectrogram) did not give good performance. Hence, we excluded them in the subsequent experiments. First, we normalised the gain of utterances. Then, we extracted the features for every frame using a 25-ms window sliding at 10-ms. 

\subsection{Generalisation using deep multi-task learning}

In this study, we built high-level representation or features of emotional states for each utterance \cite{kunHan2014dnn,lee2015high}. First the frame-level acoustic features were fed into a shared network for multiple tasks: emotion, gender, and naturalness. To examine the effect of multi-task learning in various architecture, we used DNN and LSTM. Let us denote them, DNN-MTL and LSTM-MTL, respectively. The shared network was composed of 2 hidden layers with 256 cells for LSTM-MTL and 3 hidden layers of 256 nodes for DNN-MTL. While DNN-MTL does not model temporal dynamics of emotional speech, it uses a context window with a size of 250ms \cite{kunHan2014dnn}. We did not find statistical differences in the performances between LSTM and Bi-directional LSTM (BLSTM). Moreover, later experiments indicated no more gains with higher number of layers and nodes.
Using these shared networks, we optimised the cost functions of the tasks, defined as:
\begin{equation} \label{eq:mtl}
  \epsilon_{total} = \epsilon_{main} + \displaystyle\sum_{i=1}^{N} \lambda_{i} * \epsilon_{{sub}_{i}}
\end{equation}
where $\epsilon$ is a cost function, $\lambda_{i}$ is a non-negative weight for subtask, and N is the total number of subtasks. Since we do not have pre-knowledge about which task contributes more to the performance of the main task, we empirically set the same weight ($.1$) for the cost of gender and naturalness.

\begin{figure}[!b]
\begin{minipage}[b]{1.0\linewidth}
  \centering
  \centerline{\includegraphics[width=7.8cm]{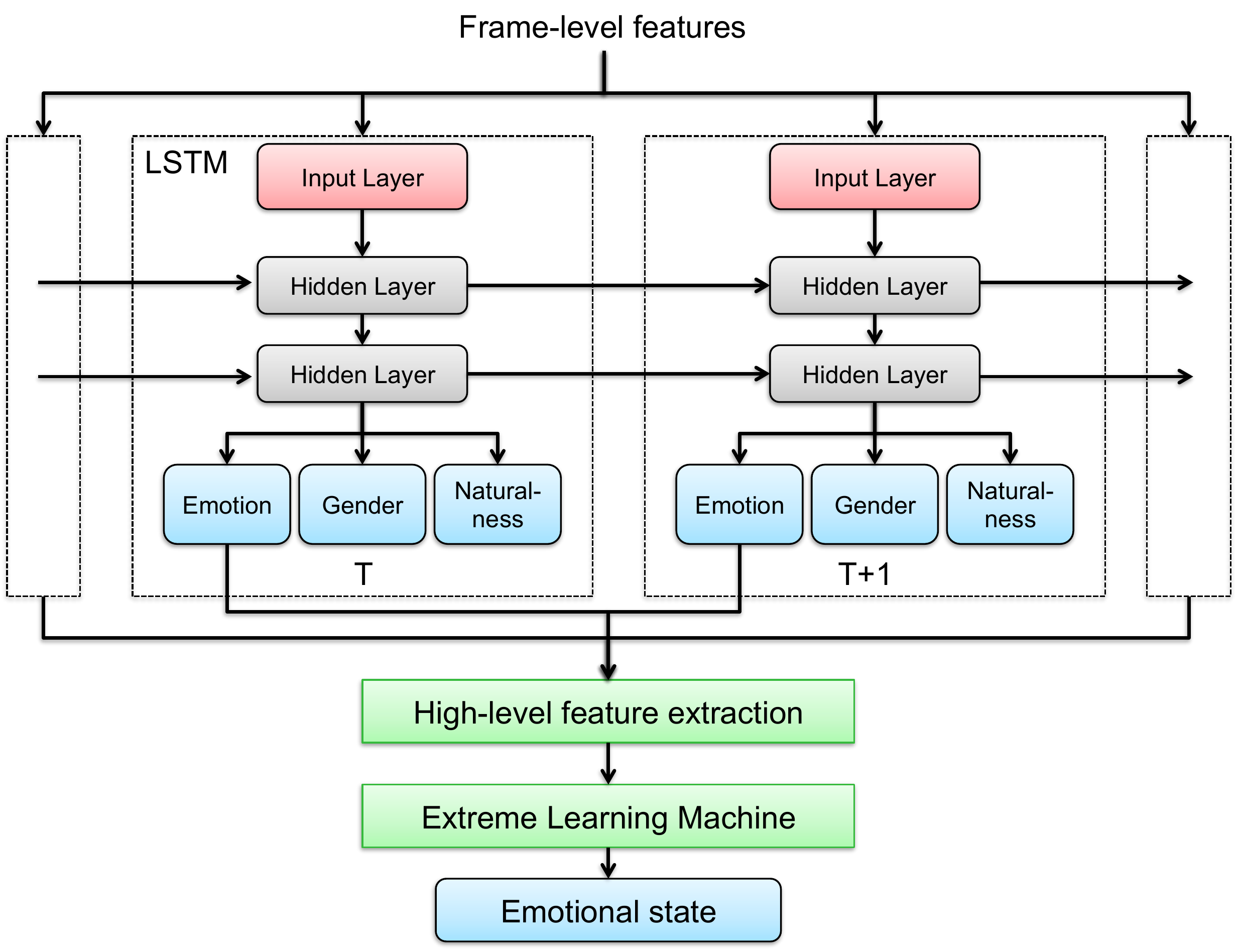}}
\end{minipage}
\caption{Block diagram of LSTM-MTL, T = time}
\label{fig:topology}
\end{figure}
Next, to build the high-level features, statistical functionals were applied to the sequential outputs of softmax layers. We used the same 4 functionals (e.g. min, max, mean, and etc.) proposed in \cite{kunHan2014dnn}. We aim to utilise the diversity of the contexts for improved generalisation of the emotion models trained and mainly examine the effect of the generalisation, not an extended representation including gender and naturalness. Hence, we discarded output layers of the contexts after training the shared network. Finally, our representation included emotional categories (4 classes), totalling 16 (4 classes x 4 functionals) high-level features for the proceeding ELM (the same setting of ELM in \cite{lee2015high}). Figure \ref{fig:topology} describes the topology of LSTM-MTL. 

\section{Experiments and Result}\label{sec:result}
\subsection{Experiment setup}
We compared DNN-MTL and LSTM-MTL to single task learning (STL) based methods: DNN-STL~\cite{kunHan2014dnn} and LSTM-STL~\cite{lee2015high}. Since we aim to examine the ability to generalise emotional speech in the diverse contexts, we composed various validation settings: \textsc{within-corpus} and \textsc{cross-corpus}. First, the \textsc{within-corpus} setting is leave-one-speaker-out-cross-validation (LOSOCV). Although this might not be an optimal setting for MTL because of the smaller sizes of the corpora, we included the results as baselines and for completeness.

In the \textsc{cross-corpus} (leave-one-corpus-out-cross-validation) condition, we test on a corpus that was not included as training data. Optimising models without any access to a testing corpus is such a challenging task. Particularly, AIBO and ENTERFACE do not cover the complete set of 4 emotional categories and show a severe unbalance in number of samples as can be seen in Table \ref{tab:corpora}. Moreover, we examine the contribution of each subtask to the emotion detection task. We compared the baseline (STL) and the proposed methods using either gender (GENDER-MTL) or naturalness (NATURALNESS-MTL), and both of them (ALL-MTL) as subtasks. In all conditions, we used 10\% of training data for optimising parameters and excluded them from training data.

As a common setting, we utilised stochastic optimisation with a mini-batch of 128 samples, Adam method~\cite{kingma2014adam}, and a fixed learning rate of $3 \cdot 10^{-3}$. We used categorical cross-entropy for the cost function. To prevent over-fitting, we used dropout~\cite{srivastava2014dropout} with $p =.5$ and early-stopping \cite{prechelt1998automatic}. 
As an evaluation matrix, we used Unweighted Accuracy (UA) to consider the unbalanced number of samples between classes. Lastly, we used Wilcoxon signed-rank paired test ($0.95$ of confidence level)~\cite{gibbons2011nonparametric} to see statistical significance of gains of DNN-MTL and LSTM-MTL over baselines. 

\subsection{Result}

Table \ref{tab:withincorpus} summarised the results of \textsc{within-corpus} experiments. Overall, LSTM-MTL ($.706$) outperformed DNN-STL ($.629$) and LSTM-STL ($.689$). 
There was no significant gains for EMODB, ENTERFACE, and LDC that have smaller number of samples compared to those of AIBO and IEMOCAP. For AIBO and IEMOCAP, LSTM-MTL showed significant gains ($p < .05$) over LSTM-STL. In short, MTL could be effective for training a single corpus if the size is sufficiently large. 
\begin{table}[!t]
\footnotesize{
\centering
\caption{Unweighted accuracy of within-corpus experiments; Mean value of UA over corpora (M) (\textbf{bold fonts for significant gains compared to baseline})}\label{tab:withincorpus}
\begin{tabular}{|p{0.2cm}|p{1.4cm}|p{1.5cm}|p{1.4cm}|p{1.5cm}|}
\hline
ID  & \multicolumn{2}{|c|}{Baseline} & \multicolumn{2}{|c|}{Proposed} \\ \cline{2-5}
  & DNN-STL & LSTM-STL      & DNN-MTL   & LSTM-MTL \\
\hline
\hline
A   & $.405\pm.03$ & $.489\pm.06$ & $.409\pm.11$      & $\mathbf{.520\pm.07}$ \\
I & $.432\pm.12$ & $.529\pm.14$ & $\mathbf{.461\pm.04}$ & $\mathbf{.569\pm.13}$ \\
E & $.888\pm.07$ & $.921\pm.10$ & $.842\pm.11$      & $.925\pm.08$ \\
F & $.867\pm.10$ & $.953\pm.15$ & $.860\pm.12$      & $.953\pm.15$ \\
L & $.554\pm.12$ & $.552\pm.15$ & $.537\pm.13$      & $.564\pm.15$ \\
\hline
\hline
M   & $.629$ &  $.689$    & $.622$ & $\mathbf{.706}$ \\
\hline
\end{tabular}
}
\end{table}

\begin{table*}[!tp]
\centering
\footnotesize{
\caption{Unweighted accuracy of emotion recognition in cross-corpus experiments (\textbf{bold fonts for gains compared to baseline})}\label{tab:cross-corpus}
\begin{tabular}{|p{1.8cm}|p{0.5cm}|p{1.2cm}|p{1.2cm}|p{1.2cm}|p{1.2cm}|p{1.2cm}|p{1.2cm}|p{1.2cm}|p{1.2cm}|}
\hline
\multicolumn{2}{|c|}{Corpus ID} & \multicolumn{2}{|c|}{Baseline (STL)} & \multicolumn{6}{|c|}{Proposed (MTL)} \\ \cline{5-10}
\multicolumn{2}{|c|}{}  & \multicolumn{2}{|c|}{}   & \multicolumn{2}{|c|}{ALL-MTL} & \multicolumn{2}{|c|}{Gender-MTL} & \multicolumn{2}{|c|}{Naturalness-MTL}\\
\hline
Train & Test & DNN-STL  & LSTM-STL      & DNN-MTL & LSTM-MTL & DNN-MTL & LSTM-MTL & DNN-MTL & LSTM-MTL\\
\hline
\hline
\{E,F,L,IN,IA\} & A &   $.310$ &  $.355$ &  $\mathbf{.351}$ & $.352$  & $.308$ &  $.338$ &  $\mathbf{.401}$ & $.326$ \\
\{A,F,L,IN,IA\} & E &   $.359$ &  $.288$ &  $\mathbf{.388}$ & $\mathbf{.425}$ & $\mathbf{.392}$ & $\mathbf{.384}$ & $.286$ &  $.267$ \\
\{A,E,L,IN,IA\} & F &   $.306$ &  $.332$ &  $\mathbf{.327}$ & $\mathbf{.351}$ & $.300$ &  $.318$ &  $\mathbf{.337}$ & $.326$ \\
\{A,E,F,IN,IA\} & L &   $.239$ &  $.251$ &  $\mathbf{.347}$ & $\mathbf{.267}$ & $\mathbf{.251}$ & $.251$ &  $\mathbf{.249}$ & $.244$ \\
\{A,E,F,L,IA\}  & IN  & $.361$ &  $.455$ &  $\mathbf{.481}$ & $\mathbf{.484}$ & $.331$ &  $.397$ &  $\mathbf{.488}$ & $\mathbf{.464}$ \\
\{A,E,F,L,IN\}  & IA  & $.378$ &  $.341$ &  $\mathbf{.498}$ & $\mathbf{.464}$ & $\mathbf{.405}$ & $\mathbf{.445}$ & $.378$ &  $\mathbf{.353}$ \\

\hline
\hline
\multicolumn{2}{|c|}{Mean of natural: A,IN} & $.335$  & $.405$    & $\mathbf{.416}$ & $\mathbf{.418}$       & $.319$ &  $.367$ &  $\mathbf{.445}$ & $.395$ \\  
\multicolumn{2}{|c|}{Mean of acted: E,F,L,IA} & $.320$  & $.303$  & $\mathbf{.390}$ & $\mathbf{.377}$ & $\mathbf{.337}$ & $\mathbf{.349}$ & $.312$ &  $.298$ \\  
\hline
\hline
\multicolumn{2}{|c|}{Overall mean}  & $.325$ &  $.337$ &  $\mathbf{.399}$ & $\mathbf{.391}$ & $\mathbf{.331}$ &   $\mathbf{.355}$ &   $\mathbf{.356}$ &   $.330$ \\

\hline
\end{tabular}
}
\end{table*}

Table \ref{tab:cross-corpus} summarised the results of \textsc{cross-corpus} experiments. Since IEMOCAP has both acted and natural emotional speech, we divided it into acted (IA) and natural (IN). First, we compared Baseline (STL) and ALL-MTL that uses both gender and naturalness as auxiliary tasks. Both DNN-MTL and LSTM-MTL showed gains for all corpora (except for AIBO by LSTM-MTL) while the gains varied on the corpora and architectures. DNN-MTL outperformed DNN-STL by \textbf{11\%}, \textbf{12\%}, \textbf{12\%} for LDC (L), improvised (IN) and scripted (IA) IEMOCAP corpora, respectively. LSTM-MTL improved LSTM-STL by \textbf{13\%}, \textbf{12\%} for EMODB (E) and scripted (IA) IEMOCAP corpora. The overall mean of gains of DNN-MTL (LSTM-MTL) over DNN-STL (LSTM-STL) was $\mathbf{7.4}$ ($\mathbf{5.4}$)\%, which is statistically significant and more superior to the gains reported in the within-corpus setting. 

Next, we examined the performance of MTL using only either gender or naturalness as an auxiliary task in order to see the contribution of each task to the main task. As shown, Gender-MTL and Naturalness-MTL showed smaller overall gains compared to ALL-MTL. For some corpora (e.g. AIBO), Naturalness-MTL showed even better performance than ALL-MTL; however, hurt of generalisation was also reported depending on the corpora and architecture. Moreover, there was no significant difference between the performance of Gender-MTL and Naturalness-MTL.

Lastly, we investigated dependency of our methods on the type of testing corpora: acted and natural. DNN-MTL and LSTM-MTL obtained more overall gains from testing acted corpora. However, AIBO has a missing category and the severe unbalanced number of samples for classes (82\% of the data was neutral in Table \ref{tab:corpora}). In addition, acted corpora often do not have prototypical but diverse emotional expressions \cite{schuller2011selecting}. Hence, we do not conclude that effect of our methods is limited to only typical expressions.


\subsection{Visualisation of representations}
\begin{figure}[!tb]
\footnotesize{
\begin{minipage}[b]{0.48\linewidth}
  \centerline{\includegraphics[width=3.8cm]{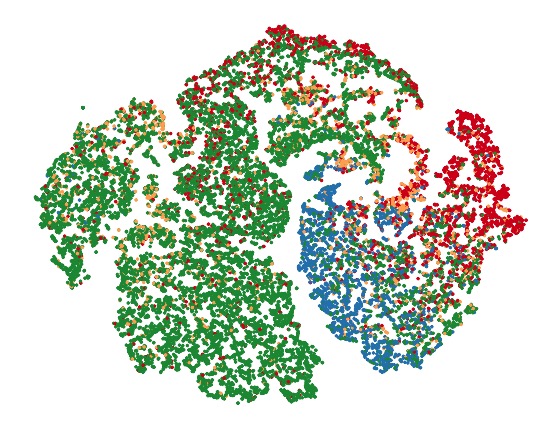}}
  \centering{(a) DNN-STL}
\end{minipage}
\begin{minipage}[b]{0.48\linewidth}
  \centerline{\includegraphics[width=3.8cm]{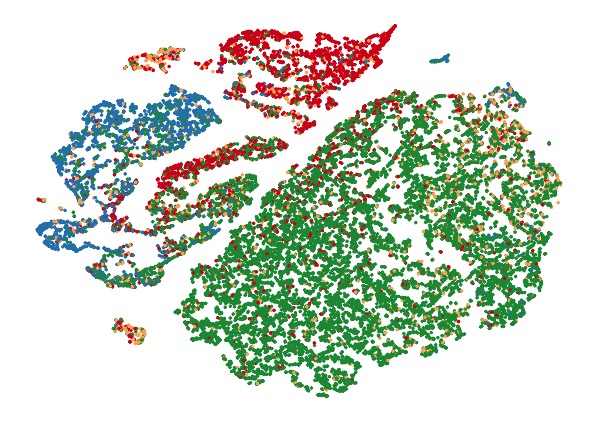}}
  \centering{(b) DNN-MTL}
\end{minipage}
\begin{minipage}[b]{0.48\linewidth}
  \centerline{\includegraphics[width=3.8cm]{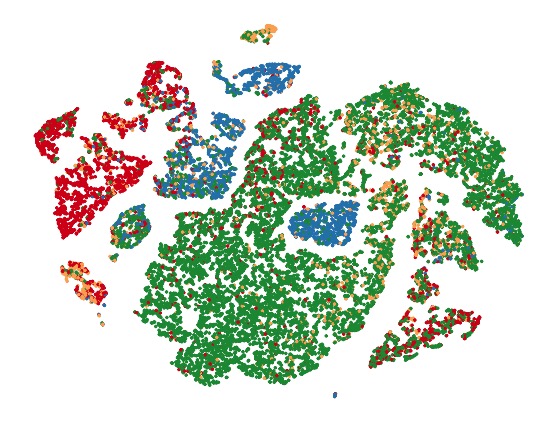}}
  \centering{(c) LSTM-STL}
\end{minipage}
\begin{minipage}[b]{0.48\linewidth}
  \centerline{\includegraphics[width=3.8cm]{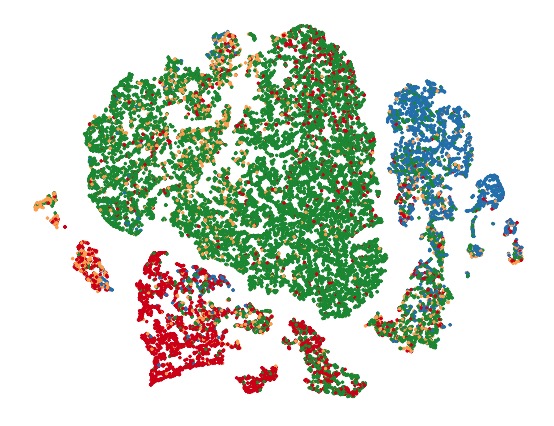}}
  \centering{(d) LSTM-MTL}
\end{minipage}
}
\caption{The result of T-SNE for high-level features of the aggregated corpora; coloured by emotional categories (green: neutral, orange: happy, blue: sad, and red: angry)}
\label{fig:tsne}
\end{figure}

We visualised high-level representations of the selected corpora to see the generalisation ability of our method proposed in a feature space. To this end, we employed t-distributed stochastic neighbour embedding (T-SNE)~\cite{maaten2008visualizing} that is a non-linear dimensionality reduction technique embedding high-dimensional data into a space of two or three dimensions. First, we aggregated the selected corpora and shuffled the utterances in a random manner. Next, we split the shuffled data into training (80\%), validation (10\%), and testing (10\% of the whole data) sets. We fed the training data into each model and optimised the parameters using the validation data. Then, the aggregated data was fed into the trained models to obtain the high-level representations as explained in Section \ref{sec:method}. In Figure \ref{fig:tsne}, we compared high level representations learnt from of DNN-STL (a), DNN-MTL (b), LSTM-STL (c), and LSTM-MTL (d). To look at benefits of these representations in a more quantitative way, we summarised their confusion matrix of the testing set in Table \ref{tab:cm}.

As shown in Figure \ref{fig:tsne}, the proposed methods, (b) and (d), showed relatively more discriminative clusters compared to those of (a) and (c). While (a) showed mixed data points of neutral, sad, and angry, (b) showed more separated clusters of these categories. In Table \ref{tab:cm} (b) showed large gains of sad and angry (\textbf{52}\% and \textbf{21}\%, respectively). When we compared (c) and (d), we could find a more separated cluster of sad that achieved a gain of \textbf{19}\%. In overall, DNN(LSTM)-MTL outperformed DNN(LSTM)-STL by large gains \textbf{17}(\textbf{6})\% and showed more generalised representations of the large aggregated corpus compared to DNN(LSTM)-STL.


\begin{table}[!tp]
\centering
\footnotesize{
\caption{Confusion matrix of the testing data and unweighted accuracy (UA)}\label{tab:cm}
\begin{tabular}{|l|l|l|l|l|l|l|l|l|}
\hline
& \multicolumn{4}{|c|}{ (a) DNN-STL} & \multicolumn{4}{|c|}{(b) DNN-MTL}\\ \cline{2-9} 
  & N & H & S & A &           N & H & S & A\\ 
  \hline
N & $.99$ & $.00$ & $.01$ & $.01$ & $.94$ & $.00$ & $.04$ & $.02$ \\
H & $.94$ & $.01$ & $.01$ & $.04$ & $.80$ & $.02$ & $.06$ & $.12$ \\
S & $.78$ & $.00$ & $.18$ & $.04$ & $.16$ & $.01$ & $\mathbf{.70}$ &  $.14$ \\
A & $.73$ & $.00$ & $.00$ & $.26$ & $.44$ & $.02$ & $.06$ & $\mathbf{.47}$ \\
\hline
UA & \multicolumn{4}{|c|}{ $.359$ } & \multicolumn{4}{|c|}{ $\mathbf{.534}$ }\\
\hline
\hline
& \multicolumn{4}{|c|}{(c) LSTM-STL} & \multicolumn{4}{|c|}{(d) LSTM-MTL}\\ \cline{2-9} 
\hline
N & $.90$ & $.03$ & $.03$ & $.05$ & $.91$ & $.01$ & $.04$ & $.04$ \\
H & $.64$ & $.12$ & $.08$ & $.16$ & $.64$ & $\mathbf{.15}$ &  $.10$ & $.10$ \\
S & $.26$ & $.03$ & $.65$ & $.07$ & $.09$ & $.03$ & $\mathbf{.84}$ &  $.04$ \\
A & $.35$ & $.03$ & $.03$ & $.60$ & $.29$ & $.06$ & $.05$ & $.61$ \\
\hline
UA & \multicolumn{4}{|c|}{ $.565$ } & \multicolumn{4}{|c|}{ $\mathbf{.628}$ }\\
\hline
\end{tabular}
}
\end{table}

\subsection{Summary and discussion}
In summary, the overall gain in the within-corpus setting was not significant. For the DNN topology, the performance slightly dropped (Table \ref{tab:withincorpus}). However, the gains for relatively larger corpora such as AIBO and IEMOCAP were still significant. Moreover, the overall gains were much larger for the cross-corpora setting (Table \ref{tab:cross-corpus}). While we could not find a significant difference between the performance of GENDER-MTL and NATURALNESS-MTL, the combination (ALL-MTL) outperformed single-task learning based methods regardless of topology. There was no hurt of generalisation by ALL-MTL. Hence, we concluded that MTL is more effective for larger corpora in a similar way of other applications \cite{caruana1998multitask}. Also, when all tasks reach best performance at approximately the same time of training, the performance of the main task could be maximised \cite{caruana1998multitask}. Hence, a control of separated learning rates should be addressed instead of the same fixed learning rate for all tasks. Moreover, complicated networks including private layers for subtasks potentially increase the performance~\cite{caruana1998multitask} as shown in other applications \cite{jou2016deep}. Lastly, we should investigate how MTL affects discrimination of specific emotional categories as future work.


\section{Conclusions}\label{sec:conclusion}
In this paper, we proposed generalisation of emotional models using large aggregated speech corpora and deep multi-task learning of commonly accessible contexts: gender and naturalness. We tackled a practical issue in the wild, that is, aggregating small but diverse corpora. To this end, we obtained high level representation of emotional speech using DNN and LSTM that utilise gender and naturalness as subtasks. We examined our method in various settings, within-corpus and cross-corpus. In the within corpus setting, the proposed method achieved significant gains for relatively larger corpora. However, in the cross-corpus setting, larger gains were reported in most of the corpora. Particularly, the combination of gender and naturalness as subtasks resulted the best gain and no hurt of generalisation regardless of its topology. Moreover, we visualised the high-level representation obtained from the proposed method in a feature space by using t-distributed stochastic neighbour embedding and found clear clusters of emotional utterances, resulting in significant gains. We concluded that our method is applicable to various topologies and corpora but potentially more effective for larger corpora. Potential improvement can be achieved using sophisticated architectures (e.g. a private network for each task) and independent learning rates for subtasks.

\section{Acknowledgements}
The research leading to the results was supported by the European Community's 7th Framework Programme under Grant agreement 610532 (SQUIRREL - Clearing Clutter Bit by Bit) and 611153 (TERESA - Telepresence Reinforcement-learning Social Agent).
\clearpage
\bibliographystyle{IEEEtran}

\bibliography{mybib}

\end{document}